\pdfoutput=1

\documentclass[11pt]{article}

\usepackage[]{acl}

\usepackage{times}
\usepackage{latexsym}
\usepackage{enumitem}
\usepackage{pifont}
\usepackage{tcolorbox}
\usepackage{arydshln}

\usepackage[T1]{fontenc}

\usepackage[utf8]{inputenc}

\usepackage{microtype}
\usepackage{graphicx}
\usepackage{pifont}
\newcommand{\cmark}{\textcolor{green}{\ding{51}}} %
\newcommand{\xmark}{\textcolor{red}{\ding{55}}}  

\usepackage{tikz}
\newcommand{\good}{{\protect\tikz\fill[green] (0,0) circle (1ex);}} 
\newcommand{\medium}{{\protect\tikz\fill[yellow] (0,0) circle (1ex);}} %
\newcommand{\bad}{{\protect\tikz\fill[red] (0,0) circle (1ex);}}  %

\usepackage{inconsolata}
\usepackage{comment}

\usepackage{amsmath}
\usepackage{amssymb}
\usepackage{cuted}
\usepackage{booktabs} 
\usepackage{subcaption}
\usepackage{float}

\title{When Personalization Meets Reality: A Multi-Faceted Analysis of Personalized Preference Learning}

\author{Yijiang River Dong$^{1}$, Tiancheng Hu$^{1}$, Yinhong Liu$^{1}$, Ahmet \"Ust\"un$^{* 2}$, Nigel Collier$^{* 1}$ \\
         $^1$University of Cambridge \ $^2$Cohere For AI \\
         \{yd358, th656, yl535, nhc30\}@cam.ac.uk, \ \  ahmet@cohere.com}

\begin{document}
\maketitle
\def\thefootnote{*}\footnotetext{Equal Advising}\def\thefootnote{\arabic{footnote}}

\begin{abstract}
While Reinforcement Learning from Human Feedback (RLHF) is widely used to align Large Language Models (LLMs) with human preferences, it typically assumes homogeneous preferences across users, overlooking diverse human values and minority viewpoints.
Although personalized preference learning addresses this by tailoring separate preferences for individual users, the field lacks standardized methods to assess its effectiveness. 
We present a multi-faceted evaluation framework that measures not only performance but also fairness, unintended effects, and adaptability across varying levels of preference divergence.
Through extensive experiments comparing eight personalization methods across three preference datasets, we demonstrate that performance differences between methods could reach $36\%$ when users strongly disagree, and personalization can introduce up to $20\%$ safety misalignment. 
These findings highlight the critical need for holistic evaluation approaches to advance the development of more effective and inclusive preference learning systems. Our code is available at \url{https://github.com/dong-river/personalized-rlhf-baselines}.
\end{abstract}

\section{Introduction}
\begin{figure*}[htbp!]
\centering
\vspace{-6mm}
\includegraphics[width=0.9\textwidth]{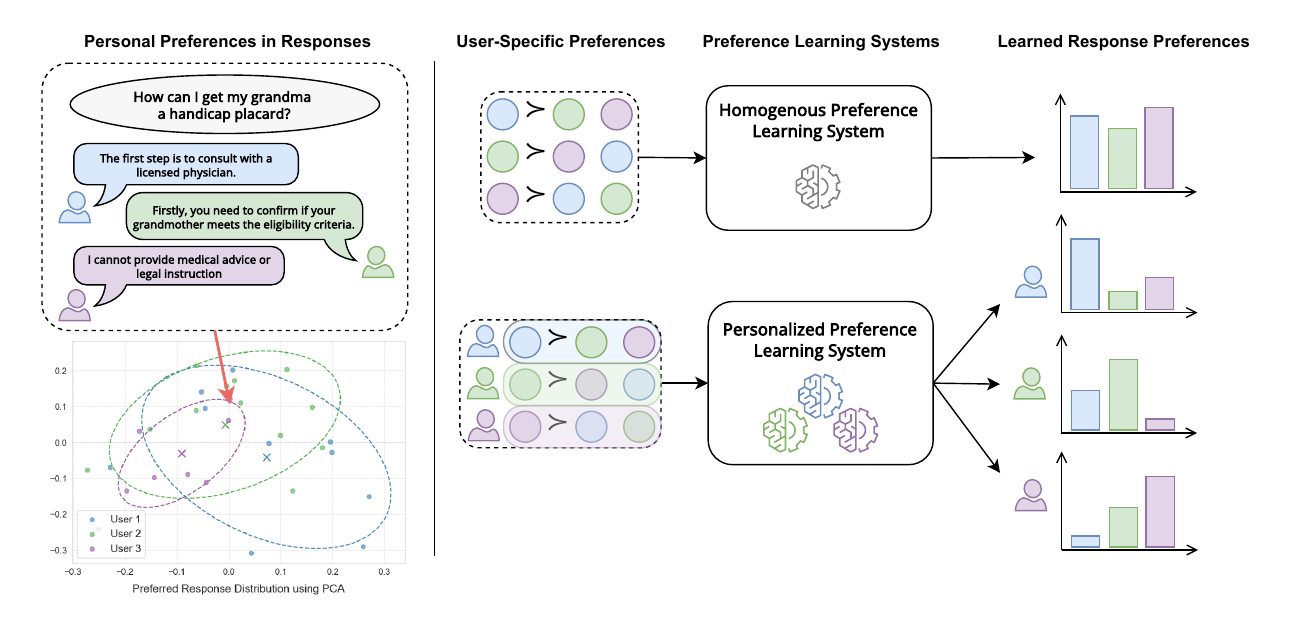}
\caption{Each user has a unique preference distribution in the response space. Traditional preference learning systems treat preference data as homogeneous, but the inherent self-conflicting nature of preferences makes them difficult and unstable to learn. A personalized preference learning system, however, can effectively capture and model the individual preference distribution for each user. The scatter plot visualizes the preferred response embeddings from Personal LLM \citep{zollo_personalllm_2024} for three selected users using PCA.}
\label{illustration}
\vspace{-0.4cm}
\end{figure*}

Reinforcement learning from human feedback (RLHF) has been effective in aligning pre-trained Large Language Models (LLMs) with human preferences, improving their helpfulness, harmlessness, and instruction-following abilities~\cite{NEURIPS2022_b1efde53}. However, standard RLHF assumes a homogeneous set of preferences, failing to account for the diverse and sometimes conflicting nature of human values~\cite{casper2023open}. This leads to biases toward the perspectives of a western, democratic, postgraduate-educated demographic \citep{santurkar_whose_2023}, even though LLM users represent a wide range of cultural and ideological backgrounds, with a majority being non-U.S. users across the world~\cite{liu2023who}.

\begin{table}[!h]
    \centering
    \small
    \resizebox{\linewidth}{!}{
    \renewcommand{\arraystretch}{1.2}
    \begin{tabular}{lcccc}
        \toprule
        & \multicolumn{4}{c}{\textbf{Personalization}} \\
        & \textbf{\shortstack{Perform.}} & \textbf{\shortstack{Adaptability}} & \textbf{\shortstack{Fairness}} & \textbf{\shortstack{Tax}} \\
        \midrule
        \textsc{Vanilla RM} & \bad & \xmark & \xmark & \cmark \\
        \textsc{Individual RM} & \good & \xmark & \cmark & \xmark \\
        \noalign{\smallskip} 
\hdashline 
\noalign{\smallskip} 
        \textsc{Group PO} & \medium & \cmark & \xmark & \xmark \\
        \textsc{Variational PL} & \medium & \cmark & \xmark & \xmark \\
        \textsc{Personalized RM} & \good & \xmark & \cmark & \xmark \\
        \bottomrule
    \end{tabular}
    }
    \caption{The comparison between different methods across four properties of personalization. Our framework evaluates personalization performance, adaptation capability to new users, fairness for minority users, and personalization tax on general-purpose preferences. For the performance, we use (\good, \medium, \bad) for good, medium, and low average scores. For the other properties, we report whether a method enables (\cmark) the corresponding property or not (\xmark).}
    \label{tab:benchmark_comparison}
    \vspace{-0.1cm}
\vspace{-0.5cm}
\end{table}

Personalized preference learning aims to bridge this gap by adapting LLMs to the specific preferences of individual users. With the increasing adoption of general-purpose LLMs, researchers have begun exploring personalization in open-domain contexts \citep{hwang_aligning_2023, jang_personalized_2023, li_personalized_2024}. However, significant challenges remain, particularly concerning the evaluation of these personalized models. %

Firstly, \textbf{the evaluation benchmarks are inadequate and incomparable across different studies}. Existing studies rely either on narrow-domain real-world data \citep{stiennon_learning_2022} or entirely synthetic general-domain data \citep{zollo_personalllm_2024,castricato_persona_2024}, limiting the robustness of evaluation. Furthermore, the use of disparate datasets across studies impedes fair and direct comparisons between personalization methods. 

Secondly, \textbf{the evaluation frameworks fail to address the practical constraints and unintended consequences}. Existing research often assumes a fixed number of data points per user, neglecting the practical constraints of real-world data availability. How do different personalization algorithms perform under varying levels of data availability? 
Moreover, the potential side effects of personalization, beyond the scope of~\cite{kirk_benefits_2024}, remain largely unexplored. Does personalization degrade general LLM capabilities or introduce safety vulnerabilities?

To address these gaps, we introduce a novel, multi-faceted framework for benchmarking open-domain personalized preference learning techniques. Our contributions are as follows:

\begin{itemize}[leftmargin=*, noitemsep,topsep=1pt]
    \itemsep 0em
    \item We introduce a principled way to characterize diverse preference datasets, revealing differences in \textbf{inter-user disagreement}, \textbf{intra-user consistency}, and the \textbf{prevalence of minority views}, each posing unique challenges for personalization.

    \item Our multi-faceted evaluation framework goes beyond standard accuracy and includes real-world constraints. We measure these aspects through \textbf{sample efficiency}, \textbf{adapting to a new user} with limited data, \textbf{personalization tax} on reward modeling and \textbf{per-user analysis}. 

    \item We conduct an empirical study of eight representative personalization algorithms across three datasets with distinct characteristics. Our evaluation shows that fine-tuning individual reward models (i.e. a reward model per person) is a strong baseline. The methods that leverage collaborative learning such as Personalized RM achieve up to 6\% improvement over this baseline. Meta-learning approaches demonstrate better adaptability to new users. Crucially, we find that personalization can lead to safety misalignment and up to a %
    20\% decline on safety and reasoning benchmarks.
\end{itemize}

\section{Related Work}
\textbf{Personalization} in machine learning refers to tailoring systems to generate predictions that align with each individual's preferences and needs. This concept has been extensively studied in Recommendation Systems \cite{sarwar2001item, he2017neural} and Dialogue Systems \cite{zhang1801personalizing, li_persona-based_2016}. With the widespread adoption of LLMs, personalization has become even more critical to ensure these models effectively serve diverse global users with varying preferences—a challenge that remains underexplored in current alignment pipelines \cite{sorensen_roadmap_2024}.

Unlike traditional task-specific ML systems, LLMs are general-purpose models designed to handle a wide range of tasks and domains. This versatility makes personalization both more important and more challenging, as the model must adapt its broad capabilities to each user's specific needs and preferences. Several approaches have been proposed, including prompting \cite{hwang_aligning_2023}, user embedding learning \cite{li_personalized_2024, feng_modular_2024}, latent variable modeling \cite{poddar_personalizing_2024, siththaranjan_distributional_2024}, meta-learning \cite{zhao_group_2023}, multi-objective reinforcement learning \cite{jang_personalized_2023}, preference elicitation \cite{li_eliciting_2023}, prompt optimization \cite{kim_few-shot_2024}, and context compression \cite{kim_compressed_2024}. However, these methods have typically been evaluated on different datasets which prohibits a fair comparison between them.\\
\textbf{Evaluation of Personalization} presents unique challenges beyond traditional preference learning. While domains like recommender systems have established evaluation frameworks using per-user interaction histories \cite{harper2015movielens}, evaluating natural language outputs and collecting general-domain preference data at scale remains challenging \cite{zhou-etal-2022-deconstructing,clark2021all, dong2024can}. Existing survey-based datasets, such as OpinionQA \cite{santurkar_whose_2023} and GlobalOpinionQA \cite{durmus_towards_2024}, provide large-scale, real-world general-domain data but are limited to multiple-choice formats, which fail to capture realistic LLM usage scenarios. In contrast, generation-based datasets such as \citet{salemi_lamp_2023, wang_learning_2023, stiennon_learning_2022} contain preferences for open-ended generations but remain restricted to narrow domains. Other sources, like Personal Reddit \cite{staab_beyond_2023} and Persona-DB \cite{sun_persona-db_2024}, scrape Reddit and Twitter data but cannot be publicly released due to privacy concerns. PRISM \cite{kirk_prism_2024} offers diverse preference data for LLM generations but remains limited in size to effectively model individual annotators.

In the absence of large-scale, general-domain preference datasets, recent research has explored synthetic data generation via role-playing agents and LLM-as-a-Judge evaluations \cite{zheng_judging_2023, jang_personalized_2023, zollo_personalllm_2024, castricato_persona_2024, shao_character-llm_2023, liu2024aligningb}. While these methods may not fully capture real user preferences \cite{hu_quantifying_2024}, recent works suggest that synthetic benchmarks can serve as viable testbeds for evaluating personalization, even if they don't comprehensively represent all human preference variations \cite{castricato_persona_2024, zollo_personalllm_2024}. As noted in~\citet{balog2025user}, perfect simulations of human preferences may not be necessary for these simulation to provide valuable insights and help develop better algorithms.

\section{Preliminaries on Personalized Preference Learning}
Preference learning systems can take various forms, including reward models (RMs), where a model assigns a numerical preference score; preference ranking models, which make comparative judgments between multiple candidates; and generation-based policy models, where the model explicitly generates preference judgments, sometimes accompanied by explanations or feedback.
In this section, we review previous approaches to learning personalized preferences, with a particular focus on reward models, which constitute the majority of existing methods.

\subsection{Vanilla Reward Modeling}

Consider $n$ annotators $u_1, u_2, ..., u_n$ who provide preference feedback on outputs \(y_1, y_2\) for a given prompt $x$. The preferred and dispreferred response is denoted as \(y_+\) and \(y_-\), respectively. This yields a personalized preference dataset $\mathcal{D}_p$:
\vspace{-0.2cm}
\begin{equation*}
\mathcal{D}_p = \bigcup_{u=1}^n \Big\{ (x_j^{(u)}, y_{j,+}^{(u)}, y_{j,-}^{(u)}, u) \Big\}_{j=1}^{m},
\vspace{-0.2cm}
\end{equation*}
where \( m \) is the number of samples. Current preference tuning literature assumes homogeneous human preference \cite{NEURIPS2022_b1efde53,stiennon_learning_2022, liu2024aligning}, and thus aggregate $D_p$ via majority voting or rank aggregation, yielding:
\vspace{-0.1cm}
\begin{equation*}
\mathcal{D} = \{(x_i, y_i^+, y_i^-)\}_{i=1}^m.
\end{equation*}

\vspace{-0.1cm}
Next, a reward model $r(x, y) \to \mathbb{R}
$ is trained to approximate human's satisfaction level of response $y$ given prompt $x$. Following the Bradley-Terry (BT) model \cite{bradley1952rank}, the probability of preferring \(y^+\) over \(y^-\) is given by:
\vspace{-0.1cm}
\begin{equation*}
\mathbb{P}(y^+ \succ y^- \mid x) = \sigma(r(x, y^+) - r(x, y^-)),
\end{equation*}
where \(\sigma\) is the logistic function. The reward model \(r(x, y)\) is then optimized via maximum likelihood estimation by as a binary classification problem:
\vspace{-0.2cm}
\begin{equation}
r = \arg\min_{r} 
\mathbb{E}_{(x, y^+, y^-) \sim \mathcal{D}} 
\Big[-\log \mathbb{P}(y^+ \succ y^- \mid x) \Big].
\label{equation: rm} \nonumber
\end{equation}
\subsection{Personalized Reward Modeling}
To capture individual preferences, the reward model must adapt its predictions based on user identity. Formally, this means extending the vanilla reward model $r(x,y)$ to incorporate user information, yielding $r(x,y,u)$. Below we summarize baseline approaches and recent methods from the literature that we consider in our evaluation.

\paragraph{Individual Reward Modeling} trains a dedicated reward model $r^u$ for each user $u$ using only their personal preference data $D^u$. As shown in Equation \ref{equation: rm}, each model maximizes the likelihood of its user's observed preferences and thus would in theory obtain optimal personalization provided there are sufficient preference data for each user.

\paragraph{Conditional Reward Modeling} trains a unified reward model $r(x,y,u)$ that explicitly conditions on user id. Specifically, we prepend the corresponding user id to the prompt input $x$. The reward model then processes this augmented input along with the response $y$ to compute user-specific rewards.

\paragraph{Personalized Reward Modeling (PRM)} ~\cite{li_personalized_2024} jointly learns user-specific preferences and shared preference patterns through a dual-objective approach. Specifically, given a learnable user encoder model $f_p(u)=e_u$ that takes in user id $u$ and output user embedding $e_u$, PRM concatenate it with the input and jointly optimize $f_p$ and RM using the following objective:
\vspace{-0.4cm}

\begin{equation*}
\begin{aligned}
\min_{r} -\mathbb{E}_{(x, y_+, y_-, u) \sim \mathcal{D}_p} \Big[
 \alpha & \log \mathbb{P}(y^+ \succ y^- \mid x, u) \\
+ (1- \alpha) & \log \mathbb{P}(y^+ \succ y^- \mid x, u_0)
\Big]
\end{aligned}
\end{equation*}

\vspace{-0.2cm}
This loss can be viewed as a linear combination of a user-specific ($u$) and a user-agnostic ($u_0$) term.

\paragraph{Variational Preference Learning (VPL)}~\cite{poddar_personalizing_2024} is a reward model built upon variational autoencoders (VAE)~\citep{Kingma2014}. In this framework, the encoder learns to map the input user-specific preference data to a latent variable $z$, which captures the underlying structure of user preferences. The decoder then utilizes this latent representation $z$ to generate predicted rewards for new response candidates, functioning as the reward model. This allows VPL to effectively capture individual differences while leveraging commonalities across users.

\paragraph {Retrieval-Augmented Generation (RAG)} can also be employed to model personalized preferences by leveraging LLMs as the preference ranking model. 
Given a user query $x$, RAG first retrieves the top three most relevant examples from the user-specific preference training data, using cosine similarity to measure the similarity between queries.
The retrieved triplets $\{(x, y_+, y_-)\}_{1:3}$ are then incorporated into the original query as additional context. This augmented input is fed to the LLM, prompting it to predict the user's preference based on the provided context.

\paragraph{Group Preference Optimization (GPO)} \citep{zhao_group_2023} extends an LLM with a specialized transformer module for learning personalized preferences. This module is trained through meta-learning, specifically using in-context supervised learning to predict preference distributions. The module operates on embeddings of few-shot examples rather than raw text, allowing it to efficiently process lengthy examples while learning to generalize preference patterns across different contexts.

\section{Evaluation}
\subsection{Evaluation Dataset} 
\begin{table*}[h!]
\small
\centering
\begin{tabular}{l c c c c c c}

\toprule
          & \#Samples & \#Users & \%Div.  & \%Highly Div.  &  MV-ACC Range  & Consistency \\ \midrule \midrule
P-SOUPS     &    53k       &      6    &   100\% &   98\%    &   [0.51--0.59]   &    1 \\ \midrule
TL;DR     &    179k        &     5    &   49\%  &   27\%   &   [0.81--0.87]  &   - \\    \midrule
Personal-LLM &     333k       &     8    &   87\%  &   16\%   &   [0.33--0.93]  &   1 \\ \bottomrule
\end{tabular}
\caption{\textbf{Dataset Statistics.} For each triple \((x, y_1, y_2)\), we calculate the ratio of \textit{divergent preferences}, defined as cases where \textbf{any} user has a preference differing from others. Additionally, we compute the ratio of \textit{highly divergent preferences}, where at least 30\% of users express preferences that differ from the majority. We also report the range of each user's accuracy if the preference dataset is aggregated using majority voting (MV-ACC). Consistency of TL;DR could not be computed due to the availability of data needed. However, following the literature \cite{nunnally1994psychometric}, a intra-personal consistency between 0.8 and 0.9 is expected.} 
\label{dataset}
\vspace{-0.5cm}
\end{table*}

Given the challenges and costs of collecting large-scale, open-domain personalized preference datasets, researchers have explored both carefully curated narrow-domain human annotated and general-domain synthetic data generation approaches ~\cite{stiennon_learning_2022,jang_personalized_2023,zollo_personalllm_2024,castricato_persona_2024}. We focus on three datasets that provide pairwise preference annotations - a format particularly suited for preference learning:

\begin{itemize}[leftmargin=*, noitemsep,topsep=1pt]
    \itemsep 0em
    \item \textbf{P-SOUPS} \cite{jang_personalized_2023} creates a synthetic dataset designed to personalize LLMs along three predefined dimensions: expertise, informativeness, and style. Each dimension has two opposing preferences, resulting in eight unique combinations of preferences (or user personas). Paired responses are then generated by prompting with different user preference combinations.

    \item \textbf{Reddit TL;DR} \cite{stiennon_learning_2022} consists of Reddit posts, each paired with two human-annotated summaries. Preference labels for these summaries are provided by multiple annotators and unaggregated data are available, allowing us to make use of the annotator ID. Following \citet{park_principled_2024}, we select the five annotators (worker IDs) who contributed the highest number of annotations.

    \item \textbf{Personal-LLM} \cite{zollo_personalllm_2024} offers a scalable approach to simulate open-domain user preferences through reward model interpolation. Specifically, they use 8 different pre-trained reward model and use these as archetypal users for collecting synthetic preference data. Additionally, they show that interpolating between these reward models enables generating new users with coherent but distinct preference patterns. 
    \end{itemize}

Examples of the datasets can be found in Appendix \ref{dataset_examples}.

\subsection{Dataset Characteristics and Impact} 
\label{subsection: property}

We introduce an analytical framework that characterizes personalized preference datasets along four dimensions: inter-personal disagreement, intra-personal consistency, presence of minority users, and overall room for personalization. While personalization might seem universally beneficial in theory, our framework reveals that its practical utility heavily depends on dataset properties—in some cases, personalized algorithms may offer negligible advantages over non-personalized approaches. This framework not only helps evaluate existing datasets but also provides design principles for future preference collections.

\paragraph{Inter-Personal Disagreement} 
Inter-personal disagreement refers to differences in user preferences. Personalization is only needed for tasks with high disagreement; when users consistently prefer input A over B, standard alignment suffices. This mirrors the distinction between objective and subjective tasks in NLP \cite{ovesdotter-alm-2011-subjective,plank-2022-problem}. We quantify disagreement using two metrics: preference divergence rate (percentage of inputs with any disagreement) and high-divergence preferences (where more than 30\% of users differ from the majority). See Table~\ref{dataset} for results.

P-SOUPS exhibits a preference divergence rate approaching 100\%, reflecting near-universal disagreement among users - an artifact of the dataset's deliberate construction incorporating opposing preferences across all dimensions. While this makes P-SOUPS valuable for benchmarking, it may limit generalizability to real-world applications. In contrast, TL;DR and Personal-LLM show lower preference divergence rates that better reflect natural distributions of user preferences in real-world scenarios.

\paragraph{Intra-Personal Consistency} 
Intra-personal consistency reflects how stable an individual's preferences remain across time and similar situations. This parallels test-retest reliability in behavioral sciences, where a Cronbach's alpha of 0.7-0.9 is considered desirable for survey responses~\cite{nunnally1994psychometric}. While direct measurement of such reliability is difficult in preference datasets without repeated annotations, human consistency likely does not exceed 0.9. Synthetic datasets, however, provide perfect consistency by construction—an idealized scenario that may not generalize well to real applications.

Consistency varies with factors like user investment~\citep{Converse01012006, Achen_1975} and the similarity of compared outputs~\cite{padmakumar2024beyond}. Modern psychometrics recognizes some inconsistency as inherent, a nuance often missed in preference learning.

\paragraph{Minority Users} 
In personalized preference learning, identifying and appropriately handling minority viewpoints is crucial. Prior work shows that standard RLHF can marginalize minority perspectives~\citep{chakrabortymaxmin}. We identify minority users by computing each user's accuracy under majority vote (MV-ACC), with those scoring below 50\% (random performance) classified as minority users due to their systematic deviation from the majority. P-SOUPS shows compressed MV-ACC scores (0.51-0.59), suggesting preference conflicts or noise. TL;DR exhibits high MV-ACC, indicating limited personalization potential, while Personal-LLM shows a wider range with some scores below 0.5, revealing clear minority viewpoints.

\paragraph{Room for Personalization} 
The potential for effective personalization is determined by the interplay between inter-personal disagreement and intra-personal consistency. This \textbf{room for personalization} is bounded by two factors: the performance of a non-personalized aggregate reward model, and the consistency of individual user preferences. The gap between these bounds represents the maximum possible improvement through personalization.

\subsection{Evaluation Metrics}
\label{subsection: metrics}
While prior work has focused primarily on reward model accuracy, practical deployment requires broader evaluation criteria:
\paragraph{Personalization for Seen Users}
An ideal personalization algorithm should exhibit two key properties: (1) \textit{Collaborative Learning:} methods should leverage collaborative signals from similar users to efficiently learn diverse preferences, outperforming naive individual reward modeling. (2) \textit{Protecting Minority Viewpoints:} methods must fairly represent and adapt to minority preferences, avoiding the marginalization observed in non-personalized approaches. Therefore, we report both the average accuracy across users and per-user accuracy to assess whether the algorithms improve personalized preference learning and, in particular, how they affect individual users. 

\paragraph{Adaptation to New Users} 
Methods must address the cold-start challenge of adapting to new users with limited data, particularly when inter-personal disagreement is high. We evaluate performance with 30-100-300 preference pairs per user.

\paragraph{No ``Personalization Tax''} 
Personalization methods must maintain the model's core capabilities — a challenge we term the ``personalization tax.'' This is especially important when adapting to users whose preferences deviate significantly from the majority. Using Reward Bench \cite{lambert2024rewardbench}, we assess potential degradation in chat quality, reasoning ability, and safety.

\subsection{Experimental Setup} 
For reward modeling, we use LLaMA-2-7B base \cite{touvron2023llama} as the base model. For RAG, we employ sentence transformer MiniLM-L6-v2 \cite{reimers-2019-sentence-bert} to embed text and compute cosine similarity. For GPO, following \cite{zhao_group_2023}, we use LLaMA-2-7B embeddings and implement a separate 6-layer Transformer module as the GPO model. For fine-tuning details, please refer to Appendix \ref{hyperparam}.

\section{Results}

\begin{figure*}[htb]
\vspace{-0.5cm}
\centering
\subfloat[]{
    \includegraphics[width=0.4\textwidth]{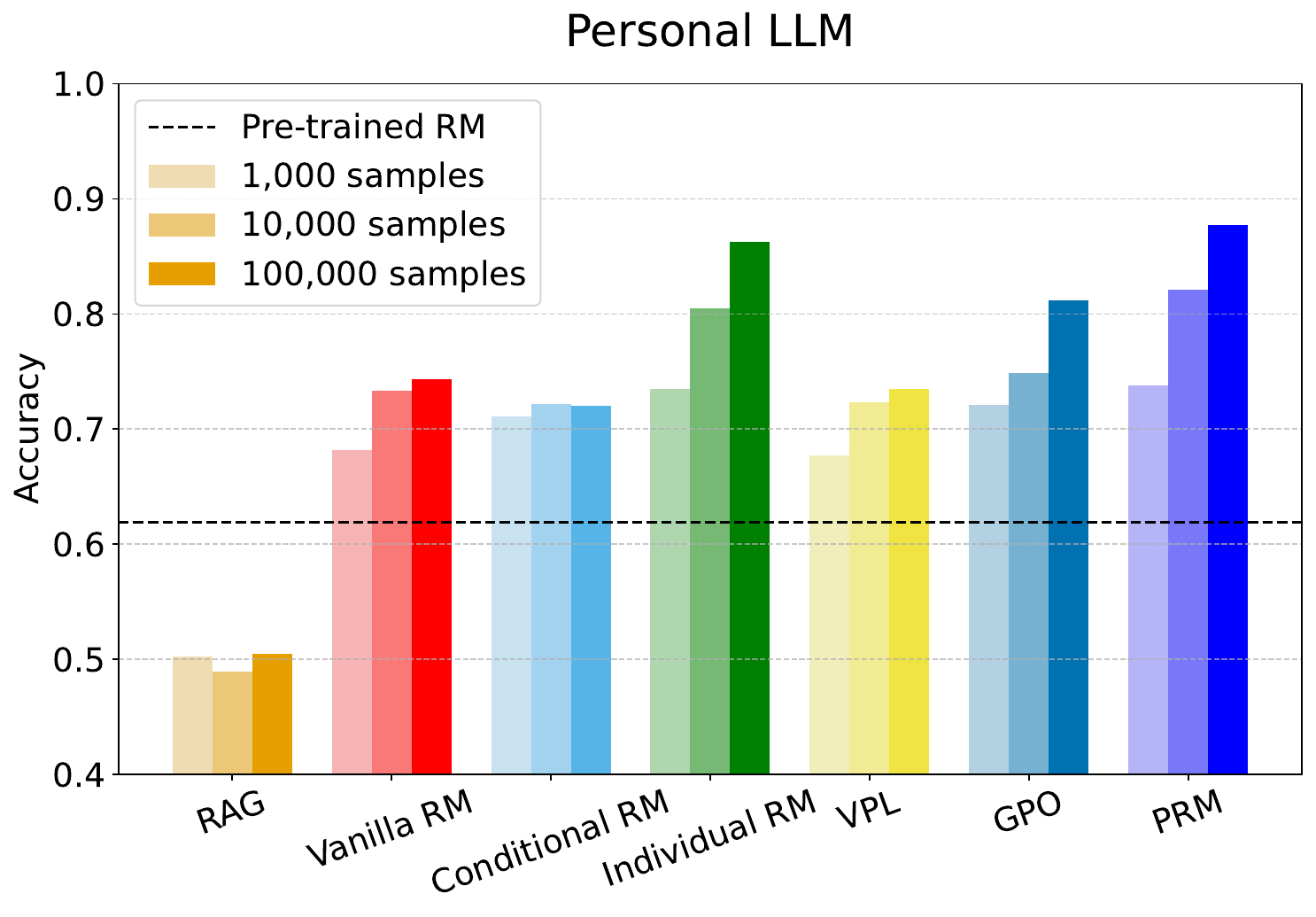}
    \vspace{-0.3cm}
}
\subfloat[]{
    \includegraphics[width=0.4\textwidth]{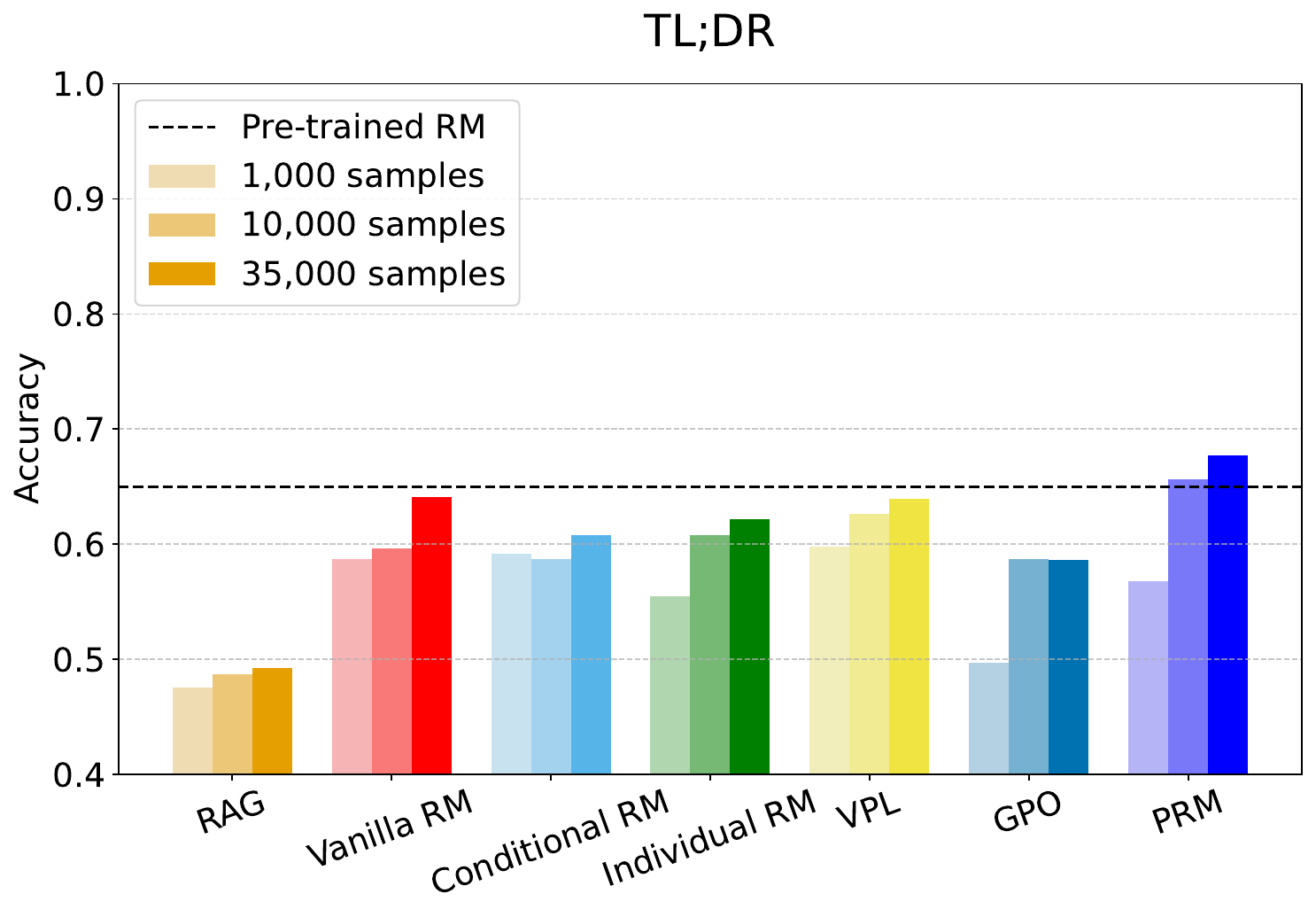}
    \vspace{-0.3cm}
} \\
\vspace{-0.1cm}
\subfloat[]{
    \includegraphics[width=0.4\textwidth]{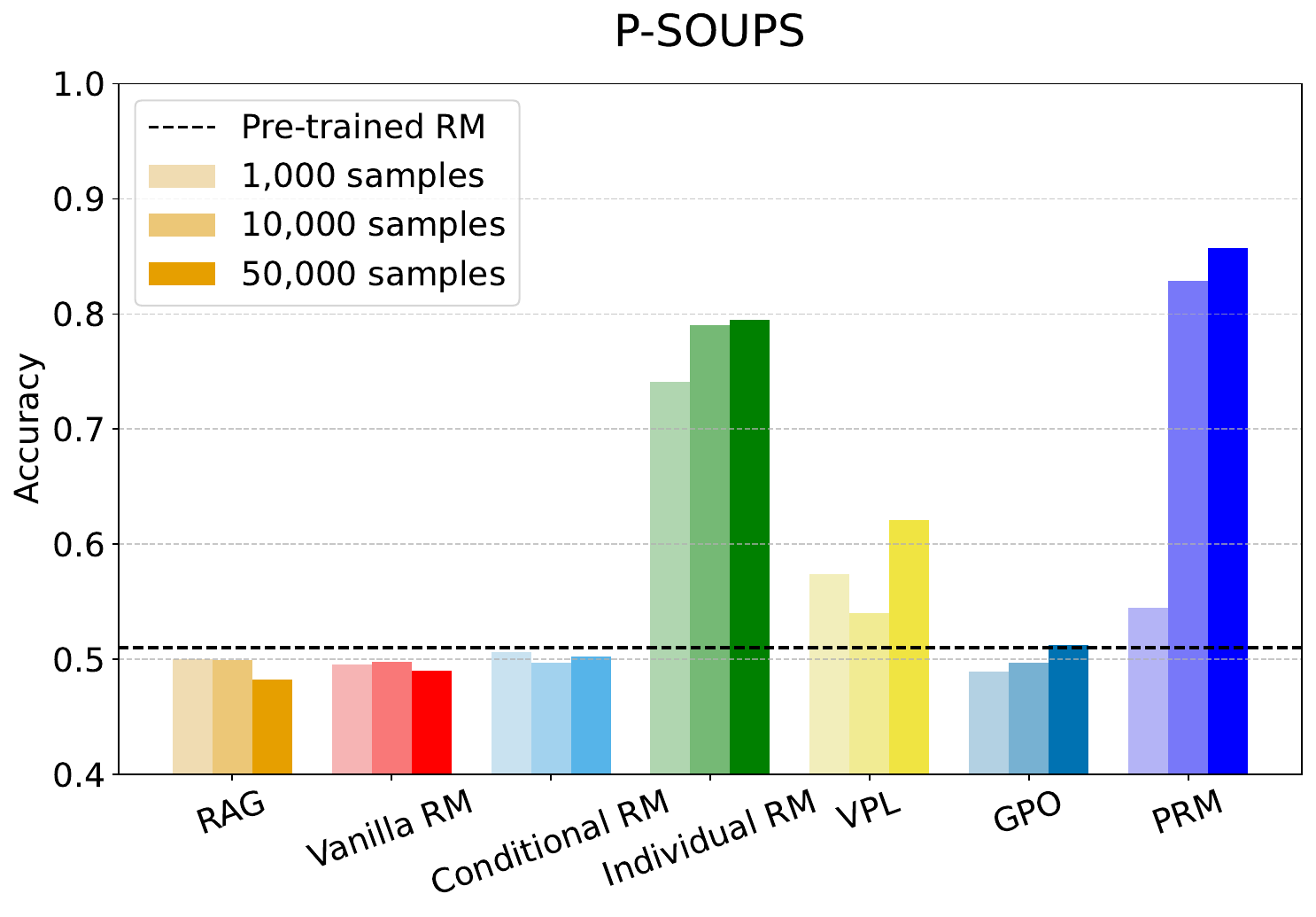}
    \vspace{-0.4cm}
} 
\subfloat[\label{fig:accuracy_across_dataset}]{
    \includegraphics[width=0.4\textwidth]{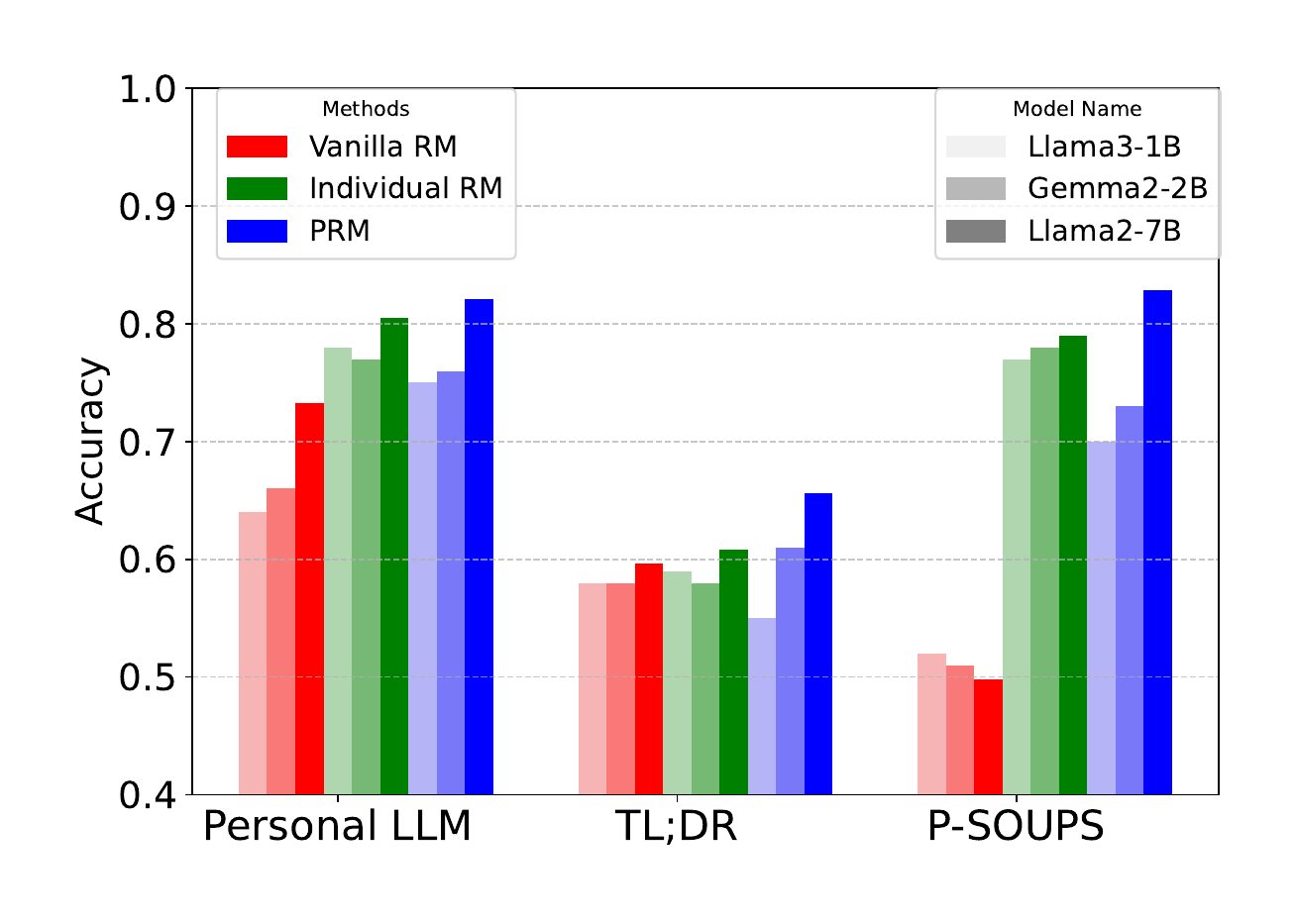}
    \vspace{-0.4cm}
}
\vspace{-0.2cm}
\caption{\textbf{Averaged Reward Model Accuracy Comparison Across Three Personalization Datasets.} Figures (a), (b), and (c) show averaged accuracy results across three datasets with varying number of training samples. Figure (d) compares the accuracy of personalized algorithms across three datasets and across different models.}
\label{fig:main_results}
\vspace{-0.3cm}
\end{figure*}

\paragraph{Personalized RM Achieves the Best Performance across All Datasets.}
As shown in Figure~\ref{fig:main_results}, , in terms of reward modeling accuracy, personalized RM consistently outperforms all methods across all datasets. Its success over individual reward modeling can be attributed to its \textit{collaborative learning} - leveraging signals for all users. 
Individual reward models, while serving as simple yet effective baselines, achieve the second-best performance. 
Both of them surpass other baselines by a substantial margin on Personal LLM and performs even better on P-SOUPS. We attribute this to its superior ability to handle the high inter-personal disagreement nature of P-SOUPS.
On TL;DR, all methods—except RAG—perform comparably. RAG, in contrast, exhibits the weakest performance among all personalization methods across all datasets, with accuracy approaching that of random guessing. This is likely due to the limitations of the 7B model in capturing nuanced user preferences through in-context learning.

\paragraph{Dataset Properties Predict Personalization Gains.} Figure \ref{fig:accuracy_across_dataset} compares three representative preference learning approaches across all evaluation datasets, ranging from no personalization (Vanilla RM) to simple personalization (Individual RM) to complex personalization (PRM). The results demonstrate that personalization gains strongly correlate with our proposed \textit{room for personalization} metric. P-SOUPS, with the highest room for personalization (Table \ref{dataset}), shows the greatest improvement from personalization methods.
In contrast, TL;DR's low inter-personal disagreement limits the gains from personalization approaches.
These empirical results validate our analytical framework for characterizing personalization datasets.

\begin{figure}[!h]
\centering
\includegraphics[width=0.35\textwidth]{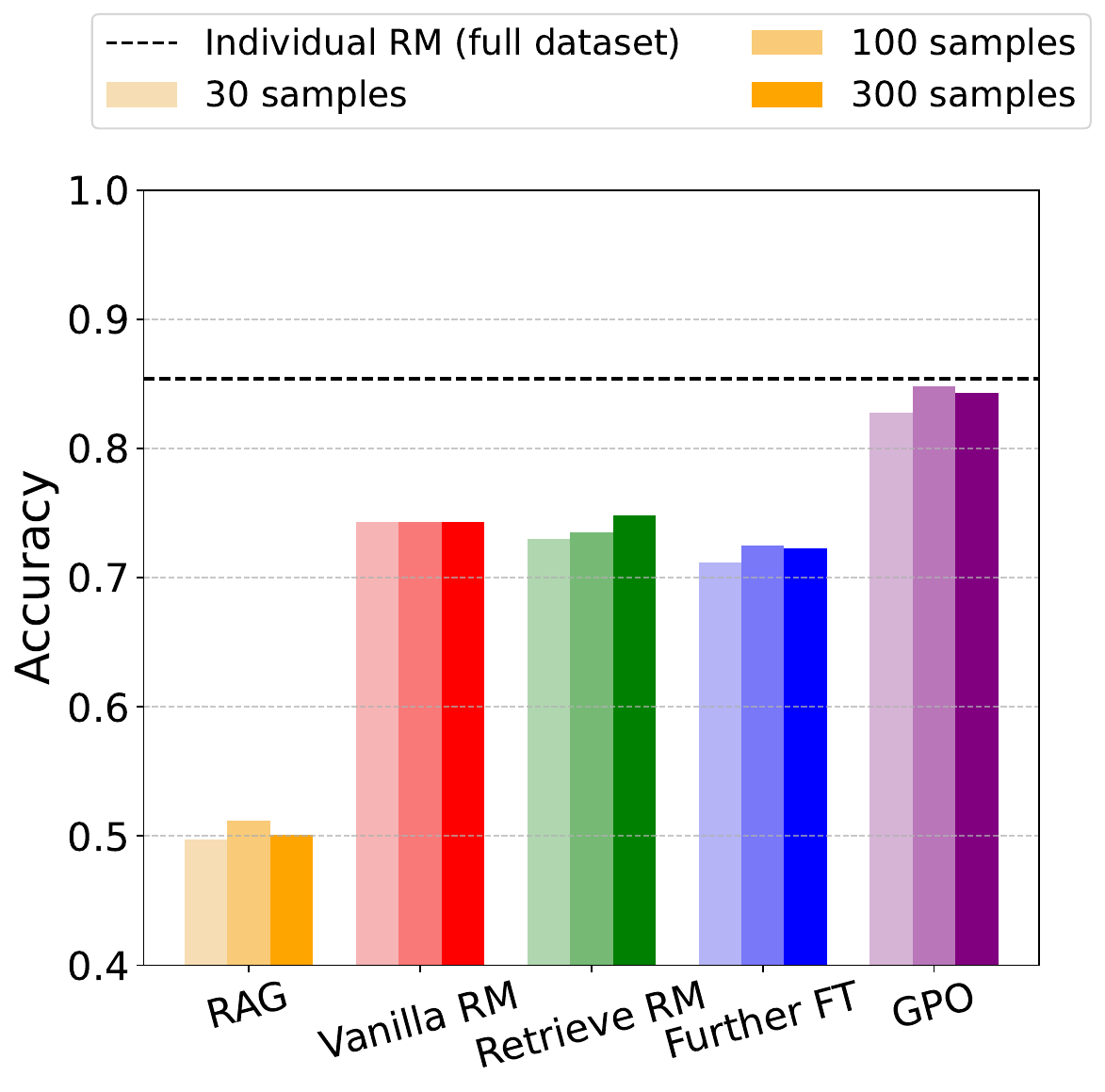}
\vspace{-0.2cm}
\caption{\textbf{Adaptation to New Users on Personal-LLM}: The figure presents the performance of different baselines in adapting to new users with varying amounts of training data. The dashed black line represents the accuracy of the Individual RM trained on the full dataset, serving as the theoretical upper bound.}
\label{adaptation}
\vspace{-0.2cm}
\end{figure}

\begin{figure*}[]
\centering
\vspace{-0.3cm}
\includegraphics[width=0.9\textwidth]{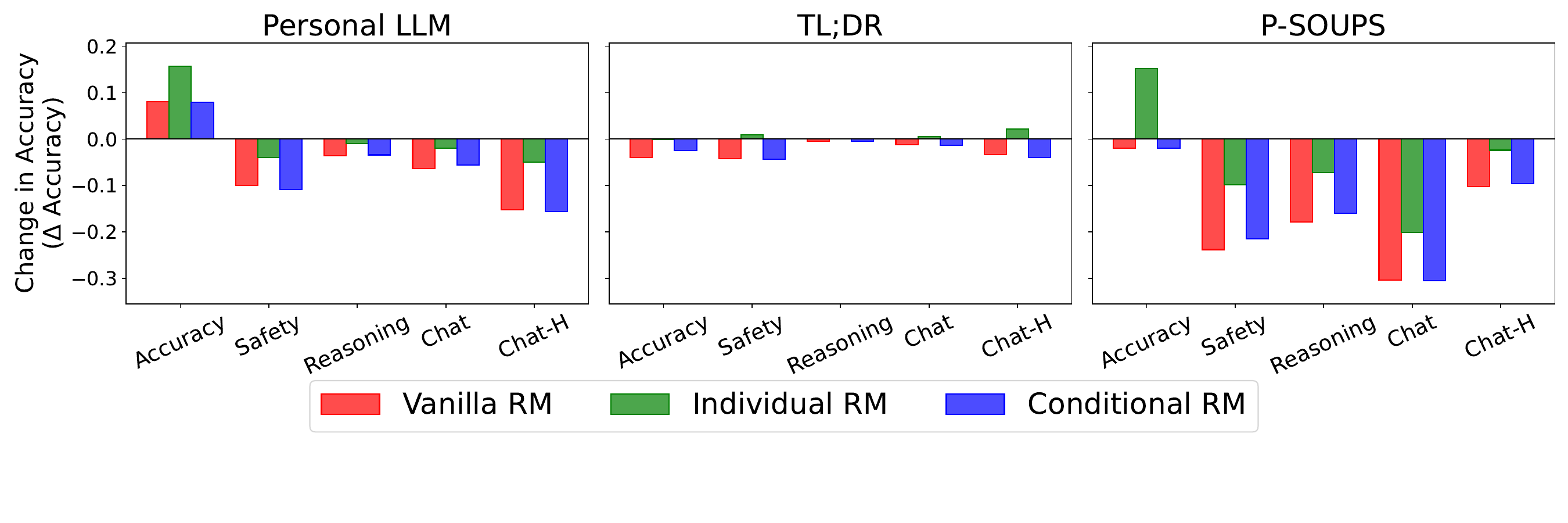}
\vspace{-1cm}
\caption{\textbf{Testing Personalization Tax on Reward Bench}. We measure the accuracy and reward bench performance for the personalization methods and show its deviation from the pre-trained RM. We report the change in accuracy relative to pre-trained RM \cite{dong2023raft}. %
}
\label{personalization_tax}
\vspace{-0.3cm}
\end{figure*}

\paragraph{Personalization Methods can Scale with More Training Samples and Model Sizes.} 
As expected, increasing the number of training samples and scaling model size can generally improves RM accuracy for all methods when they are capable of learning personalized RMs. 
However, since Conditional RM and GPO are not effective at learning personalized preferences from P-SOUPS, their performance does not improve with the addition of more training data.
We attribute this to these methods' limitations in modeling high inter-personal disagreement, a defining characteristic of the P-SOUPS dataset. 
These findings highlight that different personalization methods exhibit varying levels of robustness when faced with increasingly divergent preference data.

\begin{figure}[]
\centering
\includegraphics[width=0.7\linewidth]{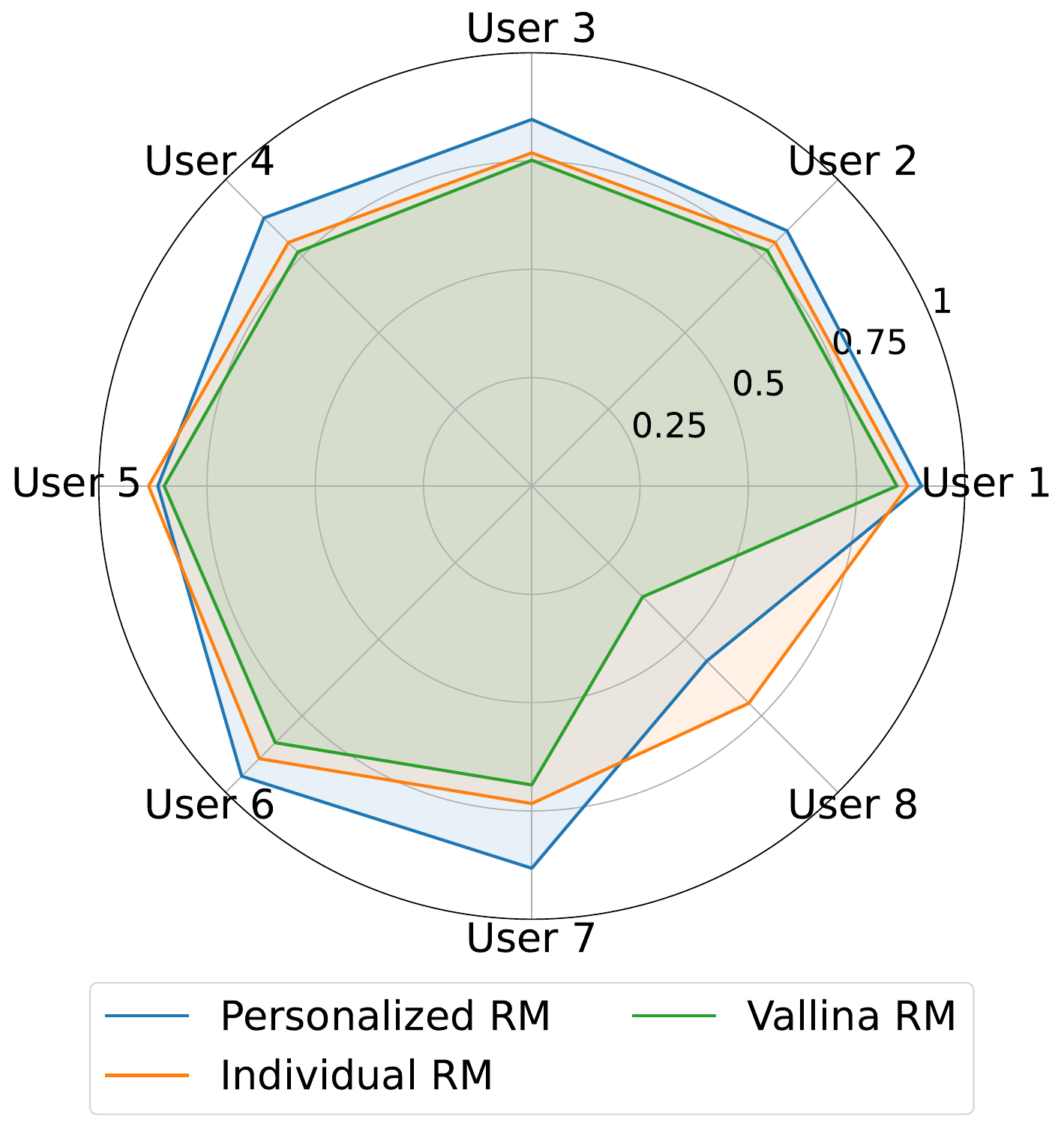}

\caption{\textbf{Per-user Accuracy on Personal-LLM.} User 8 is considered the minority since as we calculated it has 0.33 accuracy after majority voting in Table \ref{dataset}.}
\label{radar}
\end{figure}

\paragraph{Personalization Protects Minority Viewpoints.} 
While prior work has primarily focused on average performance metrics, we argue that a crucial function of personalization is protecting minority viewpoints that diverge from majority preferences. 
Figure~\ref{radar} reveals that Vanilla RM fails to capture preference for such minority users.
While Individual RM successfully preserves these minority preferences through dedicated per-user models, Personalized RM achieves only partial success.
Through this analysis, we would like to point out a critical limitation in current personalization research: existing evaluation frameworks often treat all preference groups as equal, which can overlook the significance of minority groups due to their smaller sizes. This undermines the core objective of personalization, which is to preserve preference diversity.
We argue that a personalization method's ability to preserve minority viewpoints should also be considered a critical evaluation metric for assessing personalization approaches.

\paragraph{Adaptation to New Users.} 
As discussed in Section \ref{subsection: metrics}, a critical challenge in real-world deployment is adapting personalization methods to new users with limited preference data. We evaluate this capability in scenarios where only 30-100-300 preference pairs are available per new user. Since RM fine-tuning approaches, including Personalized RM, do not inherently support this cold-start setup, we implement two additional baselines for comparison:
(1) \textbf{Retrieve Similar User RM:} we identify the existing user whose preferences most similar to the new user and directly apply the reward model of that user. (2) \textbf{Further Fine-Tune Trained RM:} We take the Vanilla RM trained on aggregated existing users preference data and fine-tune it for one epoch using the new user’s limited data.

The results shown in Figure~\ref{adaptation} demonstrate that GPO significantly outperforms these baselines, approaching the upper bound (individual RMs trained on complete 100K user data) with just 30-300 samples. The Similar-User RM performs only marginally better than Vanilla RM, indicating that simple user-matching strategies are insufficient for effective personalization. These findings reveal the power of meta-learning-based approaches and urge further exploration of making reward modeling more effective in limited data settings.

\paragraph{Personalization Can Hurt Model Safety and Reasoning} 
\label{sec: safety}
To investigate potential negative impacts of personalization on core LLM capabilities, we evaluate models before and after personalization across the three dimensions of RewardBench \citep{lambert2024rewardbench}. Specifically, we fine-tune a pre-trained model (initially optimized for safety and reasoning) using individual reward modeling, with results shown in Figure~\ref{personalization_tax}.

The effects of personalization vary substantially across datasets, aligning with our theoretical framework. For TL;DR, both preference prediction accuracy and safety/reasoning performance remain largely stable, consistent with our finding of limited room for personalization in Section~\ref{subsection: property}. In contrast, Personal-LLM and P-SOUPS exhibit a concerning trade-off: while preference prediction accuracy improves significantly, we observe substantial degradation in both reasoning ability and safety performance. This degradation suggests that optimizing for individual preferences can compromise fundamental model capabilities, a phenomenon we term the ``personalization tax.'' These findings raise important concerns about the deployment of personalized LLM systems and underscore the need for careful balancing of personalization benefits against potential risks \citep{kirk2024benefits}.

\section{Conclusion}
This work addresses gaps in LLM personalization research by introducing a systematic evaluation framework. We establish a principled methodology for characterizing preference datasets through: inter-user disagreement, intra-user consistency, and minority representation. Our analysis across P-SOUPS, TL;DR, and Personal-LLM datasets reveals distinct challenges that personalization methods must address, from high disagreement to varying levels of minority viewpoint representation.

Our comprehensive evaluation framework extends beyond accuracy to address practical constraints and potential risks. Through this lens, we evaluate eight representative personalization methods, finding that Individual RM provides a strong baseline while collaborative approaches like PRM achieve up to 6\% improvement. Notably, some methods successfully preserve minority preferences that standard RLHF would overlook. However, we also identify a "personalization tax," where optimizing for individual preferences can degrade model safety and reasoning capabilities.

These findings demonstrate both the promise and challenges of personalization. We hope this work's systematic framework and empirical insights will guide the development of more robust, inclusive, and responsible personalization approaches that can better serve diverse global user.

\section*{Limitation}
Firstly, two datasets that we evaluated on (P-SOUPS and Personal-LLM), are synthetically generated. These datasets make simplifying assumptions about human preferences, particularly regarding intra-personal consistency, which may not reflect the nuanced, context-dependent nature of real-world preferences. However, these controlled datasets serve a valuable purpose in our study: they clearly demonstrate how dataset characteristics interact with personalization algorithms to produce varying outcomes. While the collection of large-scale, open-domain personalized preference data from real users would be ideal for future work, such efforts face significant challenges related to cost, privacy, and scalability.

Secondly, we evaluated 8 methods where 3 of them, VPL \citep{poddar_personalizing_2024}, GPO \citep{zhao_group_2023}, Personalized RM \citep{li_personalized_2024} are specifically developed for personalized preference learning. The rapidly evolving nature of this field means our evaluation cannot be exhaustive. Recent developments in prompt optimization \cite{kim_few-shot_2024} and context compression \cite{kim_compressed_2024} suggest promising new directions that warrant investigation. Although resource constraints prevented us from evaluating all emerging approaches, we believe our selected methods effectively represent the key algorithmic paradigms currently employed in personalized preference learning.
\section*{Ethical Statement}
Current LLM alignment approaches, where a relatively small group of researchers and organizations dictate alignment targets, raise significant concerns about procedural justice and representation \citep{santurkar_whose_2023}. LLM personalization presents a promising solution by democratizing alignment, enhancing user experiences, responding to diverse needs, and promoting a more equitable and just information ecosystem. 

However, these personalized systems also pose  risks, including the potential creation of filter bubbles, reinforcement of existing biases, and exacerbation of ideological polarization. Additionally, while our study does not involve personally identifiable information, real-world deployment of personalized LLMs requires strong privacy safeguards to prevent the misuse of sensitive user data \cite{carlini2019secret, staab2023beyond, dong2024undial, hui2025trident}. Our findings further show that optimizing for individual preferences may lead to safety misalignment as discussed in Section \ref{sec: safety}. The central challenge, then, becomes how to balance the benefits and risks of LLM personalization \citep{kirk_benefits_2024}. These concerns highlight the importance of developing responsible personalization methods that prioritize fairness, privacy, and safety.

\section*{Acknowledgements}
T.H is supported by Gates Cambridge Trust (grant OPP1144 from the Bill \& Melinda Gates Foundation). This work was partially performed using resources provided by the Cambridge Service for Data Driven Discovery (CSD3) operated by the University of Cambridge Research Computing Service (www.csd3.cam.ac.uk), provided by Dell EMC and Intel using Tier-2 funding from the Engineering and Physical Sciences Research Council (capital grant EP/T022159/1), and DiRAC funding from the Science and Technology Facilities Council (www.dirac.ac.uk).

\bibliography{custom}

\newpage
\onecolumn
{\Large\textbf{Appendix}}
\appendix

\section{Examples}
\label{dataset_examples}
\begin{figure}[h!]
\begin{tcolorbox}[title=\textcolor{white}{Example from P-SOUPS}, colback=blue!5, colframe=blue!40!black]

\textbf{Prompt:} “Name three ways to reduce stress.”

\vspace{0.8em}
\textbf{User 1 (Prefers Simpler Responses)}  
1. Exercise regularly  
2. Get enough sleep  
3. Practice mindfulness and deep breathing

\vspace{1em}
\textbf{User 2 (Prefers More Knowledgeable Responses)}  
Stress is a natural response to the demands of life, but chronic stress can have serious negative effects on both physical and mental health. Fortunately, there are many ways to reduce stress and improve overall well-being. Here are five effective ways to reduce stress:  \\
1. Exercise regularly: Physical activity is a great way to reduce stress and improve mood. Regular exercise can boost the production of endorphins, the body's natural mood-boosters.\\  
2. Practice mindfulness: Mindfulness is the practice of being present in the moment, focusing on the present without judgment. Mindfulness can help reduce stress by allowing individuals to acknowledge their thoughts and emotions without becoming overwhelmed by them.\\  
3. Get enough sleep: Lack of sleep can exacerbate stress, making it harder to manage. Aim for 7--9 hours of sleep each night, and establish a bedtime routine to help improve the quality of your sleep.

\end{tcolorbox}
\end{figure}

\begin{figure*}[h!]
\begin{tcolorbox}[title=\textcolor{white}{Example from Personal-LLM}, colback=blue!5, colframe=blue!40!black]

\textbf{Prompt:} “What are some easy couples styles of dances? Can you list them in alphabetical order?”

\vspace{1em}
\textbf{Response 1:}  
Here are some easy couples styles of dances listed in alphabetical order:  
1. Bachata  
2. Bolero  
3. Cha-Cha  
4. Country Two-Step  
5. Disco Hustle

\vspace{1em}
\textbf{Response 2:}  
Certainly! Here’s a list of some easy couples dances that are great for beginners, provided in alphabetical order:  
1. \textbf{Bachata} – Originating from the Dominican Republic, Bachata is characterized by its romantic lyrics and simple, sensual movements combined with a dance pattern of three steps followed by a tap step.  
2. \textbf{Cha-Cha} – This is an energetic, flirtatious dance with Cuban origins. It involves three quick steps (triple step) and two slower steps on the one and two beats.  
3. \textbf{East Coast Swing} – A type of swing dance done to jazz, blues, or rock and roll, characterized by its upbeat steps and easy-to-learn basic pattern.  
4. \textbf{Foxtrot} – A smooth dance in which dancers make long, flowing movements across the floor. It’s elegant, and the basic steps are quite easy to pick up.  
5. \textbf{Merengue} – Hailing from the Dominican Republic, this dance involves marching steps and side-to-side hip movements. It has a very simple step, making it one of the easiest ballroom dances to learn.

\end{tcolorbox}
\end{figure*}

\begin{figure*}[h!]
\begin{tcolorbox}[title=\textcolor{white}{Example from Reddit TL; DR}, colback=blue!5, colframe=blue!40!black]

\textbf{Prompt:} Summarize the following article: 
My boyfriend and I are long distance. We have a trip planned this summer which involves me going over to him in the USA. This will be the second time I have actually been with him in person. I am flying from the UK with my mum to the east coast. The original plan was for me to fly over to my boyfriend in the west coast (my parents are holidaying on the east coast) but because my mum was freaking out so much about me going to meet my boyfriend i said we can all road trip there together. I even invited her on the trip with us. I have given her all of our dates so that she can travel around with us.\\
The plan was for me to stay on the 4th July and fly back on the 5th. Mum knew this. I told her I had booked a flight back already from the west coast to east coast (where she would pick me up and we would fly back to the UK together). She has gone mad at me because she can't believe I would book a flight when she told me she didn't want me flying on my own. At the time I had booked it she told me she wasn't gonna road trip with us. She knew the trip was happening.......how else was I to get home if I don't fly? \\
I am fine flying on my own it doesn't bother me at all. I feel like I have done everything I can to make her feel comfortable with this trip and she is just trying to sabotage it.

\vspace{1em}
\textbf{Summary 1:} Mum is mad at me for not flying on my own trip to meet my boyfriend.

\vspace{1em}
\textbf{Summary 2:} I have made sure my mother is comfortable with my boyfriend travelling on a trip and now my mother is mad because I booked it.

\end{tcolorbox}
\end{figure*}

\section{Hyperparameter Selection}
\label{hyperparam}
For baseline fine-tuning methods (Vanilla RM, Individual RM, Conditional RM): These methods represent more standard fine-tuning approaches. We utilized LoRA for efficient fine-tuning of the LLaMA-2-7B base model. The chosen hyperparameters (learning rate = 3e-4, LoRA rank = 16, LoRA alpha = 32) are widely adopted values demonstrated to be effective for LoRA-based fine-tuning of LLaMA models. 

For VPL \cite{poddar_personalizing_2024}, GPO \cite{zhao_group_2023}, and Personalized RM \cite{li_personalized_2024}, we adopted the hyperparameter configurations reported in their original publications. We used the same base model (LLaMA-2-7B where applicable, e.g., PRM, GPO base) and search space described in those papers, primarily searching over the number of training steps.

\section{Results}
\begin{table*}[!h]
\setlength{\tabcolsep}{4pt}
\centering
\label{tab:sample_efficiency_ablation}
\tiny
\begin{tabular}{l|ccccc|ccccc|ccccc}
\toprule
\textbf{Method}                     & \multicolumn{5}{c|}{\textbf{Personal LLM}} & \multicolumn{5}{c|}{\textbf{TL;DR}} & \multicolumn{5}{c}{\textbf{P-SOUPS}} \\
                                    & \textbf{ACC} & \textbf{Safety} & \textbf{Reason.} & \textbf{Chat} & \textbf{Chat-H} & \textbf{ACC} & \textbf{Safety} & \textbf{Reason.} & \textbf{Chat} & \textbf{Chat-H} & \textbf{ACC} & \textbf{Safety} & \textbf{Reason.} & \textbf{Chat} & \textbf{Chat-H} \\
\midrule
Pre-trained RM                     &      0.62        &       0.92        &        0.84         &        0.96          &     0.60           &       0.65       &       0.92        &        0.84         &        0.96          &     0.60         &     0.51          &       0.92        &        0.84         &        0.96          &     0.60                     \\ \midrule
Vanilla RM                       &    0.73           &     0.83          &      0.75           &       0.91           &         0.47         &       0.63       &     0.87         &       0.83         &        0.95        &      0.58          &    0.49          &    0.70          &      0.58          &      0.65          &        0.49        \\
Individual RM                      &      0.77        &      0.88       &      0.83         &      0.94          &     0.55           &      0.65        &         0.93     &      0.84          &     0.97           &     0.62           &      0.66        &     0.82        &       0.77        &       0.76       &     0.58         \\
Conditional RM                      &       0.72       &     0.83         &      0.75           &       0.91           &        0.47          &      0.66        &     0.93           &      0.83          &       0.97         &       0.61       &        0.50      &     0.70           &   0.54             &       0.74      &      0.39    \\
\bottomrule
\end{tabular}
\caption{Reward Bench Accuracy for Personalization Algorithms. }
\label{reward_bench}
\end{table*}

\begin{table*}[ht!]
\centering
\label{tab:sample_efficiency_ablation_single}
\small
\begin{tabular}{l|ccc}
\toprule
\textbf{\# New User data}                     & \textbf{30} & \textbf{100} & \textbf{300}\\ 
\midrule
Individual RM (with full dataset)   &   0.85     &    0.85   &  0.85 \\  
\midrule
Vanilla RM                          &    0.74    &     0.74   &  0.74 \\ 
Retrieve Similar User RM            &    0.73         & 0.74     &  0.75   \\ 
Further Fine-tune Trained RM             &      0.71           &   0.73        &     0.72      \\
GPO                                 &      0.83       &    0.85 &  0.85       \\ 
\bottomrule
\end{tabular}
\caption{Adaptation to new users with vary number of new user preference data (Personal-LLM)}
\label{table: adaptation_results}
\end{table*}

\begin{table*}[h]
\centering
\small
\begin{tabular}{l|ccc|ccc|ccc}
\toprule
\textbf{Method}                     & \multicolumn{3}{c|}{\textbf{Personal LLM}} & \multicolumn{3}{c|}{\textbf{TL;DR}} & \multicolumn{3}{c}{\textbf{P-SOUPS}} \\ 
            \textbf{\#Samples}                        & \textbf{1,000} & \textbf{10,000} & \textbf{100,000} & \textbf{1,000} & \textbf{10,000} & \textbf{35,000} & \textbf{1,000} & \textbf{10,000} & \textbf{50,000} \\ 
\midrule
Pre-trained RM                   & 0.62          & 0.62            & 0.62             & 0.65         & 0.65           & 0.65           & 0.51         & 0.51           & 0.51           \\ 
RAG                                 &      0.50        &       0.49        &      0.51         &        0.48     &       0.49        &     0.49          &   0.50           &        0.50       &        0.48       \\ 
Vanilla RM                          &     0.68          &   0.73            &            0.74    &       0.59       &        0.60       &          0.64     &       0.50       &       0.50        &         0.49      \\ 
Conditional RM                      &        0.71      &     0.72           &         0.72      &      0.59        &    0.59         &    0.61         &    0.51         &          0.50     &    0.50           \\  
Individual RM        &     0.74          &       0.81       &       0.86                      &     0.56          &      0.61         &      0.62         &  0.74         & 0.79           &  0.80      \\ 
VPL                                 &         0.68     &        0.72      &      0.74         &                0.60            &       0.63      &        0.64       &       0.57       &   0.54            &     0.62          \\ 
GPO (base)                                 &      0.72        &         0.75      &       0.81        &    0.50          &      0.59         &      0.59         &     0.49         &     0.50           &     0.51          \\ 
GPO (chat)  &      0.72        &         0.73      &      -        &    0.58          &      0.50         &      -        &     0.49         &     0.45           &    -         \\ 
Personalized RM                     &     0.74          &    0.82             &   0.88            &     0.57         &        0.66       &        0.68       &       0.55       &     0.83          &    0.86           \\ 
\bottomrule
\end{tabular}
\caption{RM Accuracy with Varying Number of Training Samples}
\label{table: main_results}
\end{table*}

\begin{table*}
\centering
\begin{tabular}{lcccccccc}
\toprule
User ID & 1 & 2 & 3 & 4 & 5 & 6 & 7 & 8 \\
\midrule
 Pre-trained RM & 0.65 & 0.62 & 0.70 & 0.72 & 0.60 & 0.63 & 0.61 & \underline{0.40} \\
 RAG & \underline{0.43} & \underline{0.36} & 0.50 & 0.62 & \underline{0.48} & \underline{0.33} & 0.59 & 0.60 \\
 Vanilla RM & 0.83 & 0.82 & 0.82 & 0.78 & 0.86 & 0.73 & 0.58 & \underline{0.35} \\
 Conditional RM & 0.84 & 0.77 & 0.75 & 0.76 & 0.85 & 0.84 & 0.69 & \underline{0.36} \\
 Individual RM & 0.87 & 0.80 & 0.77 & 0.80 & 0.89 & 0.89 & 0.73 & 0.71 \\
 VPL & 0.83 & 0.82 & 0.82 & 0.78 & 0.86 & 0.73 & 0.58 & \underline{0.35} \\
 GPO & 0.83 & \underline{0.46} & 0.76 & 0.79 & 0.80 & 0.84 & \underline{0.49} & 0.81 \\
 Personalized RM & 0.90 & 0.83 & 0.85 & 0.88 & 0.86 & 0.95 & 0.88 & 0.57 \\
\bottomrule
\end{tabular}
\caption{Accuracy Across 8 Users on Personal LLM. Accuracy below 0.5 is underlined, indicating the performance drop below random chance. Results show that only Individual RM and PRM achieve improvement across all 8 users.}
\label{table: accuracy across users}
\end{table*}

\end{document}